\documentclass[10pt,twocolumn,letterpaper]{article}

\usepackage[pagenumbers]{cvpr} 

\newcommand\blfootnote[1]{%
  \begingroup
  \renewcommand\thefootnote{}\footnote{#1}%
  \addtocounter{footnote}{-1}%
  \endgroup
}

\usepackage[accsupp]{axessibility}  

\usepackage{pifont}
\usepackage{multirow}
\usepackage{algorithm}
\usepackage{algpseudocode}

\definecolor{cvprblue}{rgb}{0.21,0.49,0.74}
\usepackage[pagebackref,breaklinks,colorlinks,allcolors=cvprblue]{hyperref}

\def\paperID{18292} 
\def\confName{CVPR}
\def\confYear{2025}

\title{From Sparse to Dense: Camera Relocalization with Scene-Specific Detector from Feature Gaussian Splatting}

\author{
    Zhiwei Huang$^{1,2}$\quad
    Hailin Yu$^{2\dagger}$\quad
    Yichun Shentu$^{2}$\quad
    Jin Yuan$^{2}$\quad
    Guofeng Zhang$^{1\dagger}$\\[1.5mm] 
    $^{1}$State Key Lab of CAD \& CG, Zhejiang University\quad
    $^{2}$SenseTime Research
}
\begin{document}
\maketitle
\begin{abstract}
This paper presents a novel camera relocalization method, STDLoc, which leverages Feature Gaussian as scene representation. STDLoc is a full relocalization pipeline that can achieve accurate relocalization without relying on any pose prior.  Unlike previous coarse-to-fine localization methods that require image retrieval first and then feature matching, we propose a novel sparse-to-dense localization paradigm. Based on this scene representation, we introduce a novel matching-oriented Gaussian sampling strategy and a scene-specific detector to achieve efficient and robust initial pose estimation. Furthermore, based on the initial localization results, we align the query feature map to the Gaussian feature field by dense feature matching to enable accurate localization. The experiments on indoor and outdoor datasets show that STDLoc outperforms current state-of-the-art localization methods in terms of localization accuracy and recall. Our code is available on the project website: 
\href{https://zju3dv.github.io/STDLoc/}{https://zju3dv.github.io/STDLoc}.

\end{abstract}    

\blfootnote{$^{\dagger}$Corresponding authors.}
\blfootnote{This work was partially supported by NSF of China (No.62425209).}

\section{Introduction}
\label{sec:intro}

\begin{figure}[t]
  \centering
  \includegraphics[width=1\linewidth]{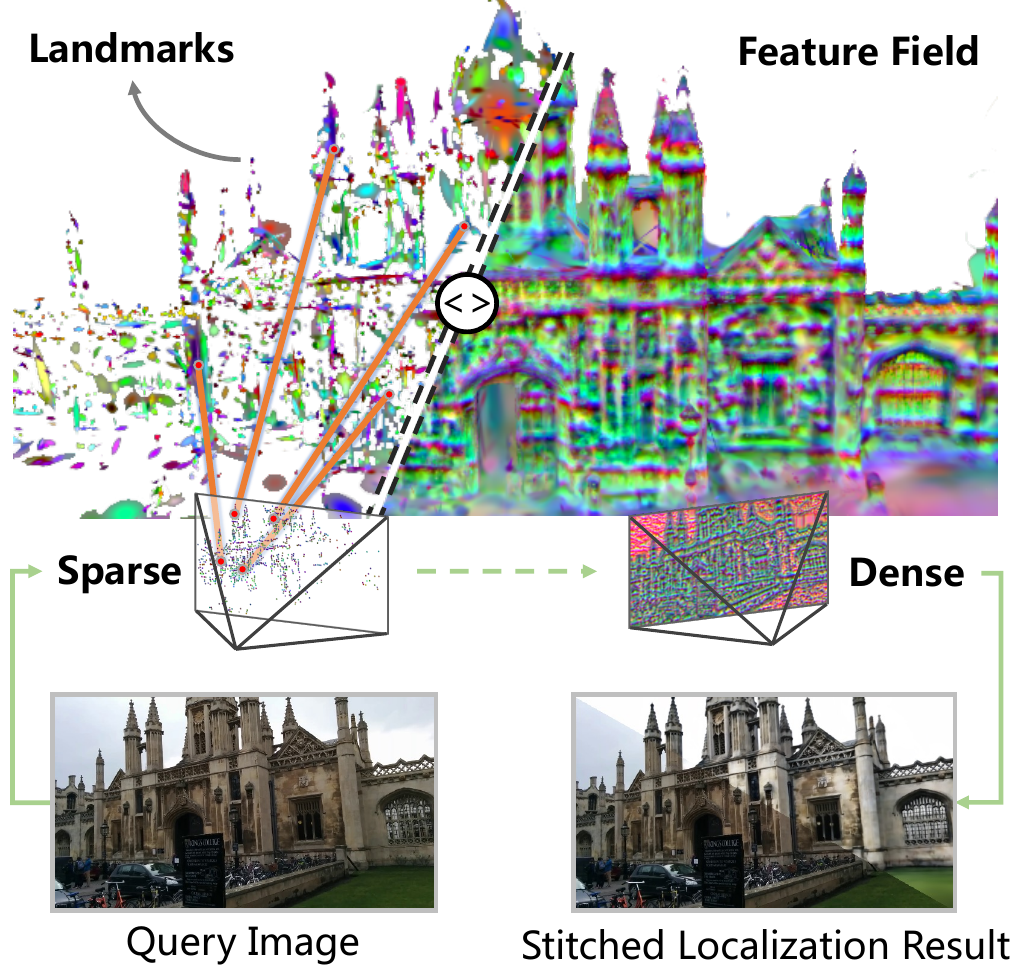}

   \caption{\textbf{STDLoc: Sparse-to-Dense Localization.} We leverage Feature Gaussian as the scene representation, which supports direct 2D-3D sparse matching on landmarks and enables the alignment of the query feature map to the feature field through dense matching.}
   \label{fig:teaser}
\end{figure}

Camera relocalization is a fundamental task in computer vision. It involves estimating the 6DoF camera pose from a query image relative to a pre-built scene map. This capability has broad applications in fields such as AR/VR, autonomous driving, and robotics.

An appropriate scene representation is critical for camera relocalization. Traditional methods \cite{li2010location, sattler2012improving,svarm2016city,liu2017efficient,taira2018inloc,sarlin2019coarse,sarlin2020superglue} leverage sparse 3D point clouds pre-reconstructed by Structure-from-Motion (SfM) \cite{schonberger2016structure} as the scene representation, with each point associated with one or more 2D descriptors \cite{lowe1999sift}. In the localization stage, reliable local features \cite{detone2018superpoint,revaud2019r2d2, potje2024xfeat,lowe1999sift} are first extracted from the query image, and these features are matched either with reference images \cite{sarlin2020superglue,sun2021loftr} or directly with the 3D point cloud \cite{sattler2012improving,sarlin2019coarse, yu2022improving,yan2023long}. Then, the 6DoF pose is estimated using the Perspective-n-Point (PnP) \cite{lepetit2009epnp} algorithm based on the established 2D-3D correspondences. This type of method, also called feature matching-based localization, can achieve high localization accuracy in rich-texture scenarios but degrade in weak-texture environments \cite{shotton2013scene} due to insufficient feature correspondences. Semi-dense \cite{sun2021loftr, giang2024learning} or dense matching \cite{edstedt2024roma} can effectively alleviate this problem, but makes SfM computation unacceptable at the mapping stage. Mesh can provide dense depth, but the localization accuracy is seriously affected by artifacts \cite{panek2022meshloc, panek2023visual}.

Many works have explored encoding scene information into a deep neural network. They train a neural network to regress absolute poses directly (APR) \cite{kendall2015posenet, chen2021direct} or scene coordinates (SCR) \cite{shotton2013scene, brachmann2023ace}. The APR method generally has limited accuracy and poor generalization to unseen views \cite{brachmann2021limits}. Due to the regression of dense scene coordinates, the SCR method is more accurate than the feature matching-based localization method in indoor weak-texture scenes \cite{brachmann2021visual}. However, since it directly encodes the scene information into the network weights, it cannot dynamically adjust the network capacity according to the size of the target scene. Therefore, its accuracy in large outdoor scenes is relatively limited \cite{sarlin2021back}.

Recently, implicit scene representations represented by NeRF \cite{mildenhall2021nerf} and explicit scene representations represented by 3D Gaussian Splatting (GS) \cite{kerbl20233d} have demonstrated photographic novel view synthesis ability. Therefore, some works explore leveraging NeRF \cite{yen2021inerf, zhou2024nerfmatch} and Gaussian \cite{Bortolon20246dgs, sun2023icomma} as scene representation to design localization algorithms, showing promising results. However, most current methods do not have a full pipeline. Some methods \cite{chen2022dfnet, moreau2022lens} utilize the novel view synthesis capability of NeRF or GS for data augmentation. 
Another direction of research focuses on pose refinement through iterative optimization. Although some of the current methods \cite{yen2021inerf, sun2023icomma, botashev2024gsloc} achieve high accuracy by minimizing the photometric error, which is unsuitable for environments with illumination changes. Moreover, these approaches rely on external methods to provide initial pose estimates. 

To address the abovementioned limitations, we propose STDLoc, a novel camera relocalization method based on Gaussian Splatting, as shown in \cref{fig:teaser}. Inspired by Feature 3DGS \cite{zhou2024feature}, we leverage feature field distillation to extend the functionality of the original GS so that it can represent not only the radiance field but also the feature field, which we call Feature Gaussian and use as the scene representation for localization. 
Based on this representation, we make the following contributions. 
(1) We propose a novel sparse-to-dense camera relocalization pipeline that leverages Feature Gaussian as the scene representation.  The proposed localization method uses sparse features to obtain the initial pose and dense features for pose refinement, which enables accurate camera relocalization. Instead of performing image retrieval and then feature matching, this pipeline achieves a new coarse-to-fine localization paradigm.
(2) We introduce a novel matching-oriented sampling strategy to address the challenge of selecting landmarks from millions of Gaussians. This strategy significantly reduces the number of Gaussians, selecting only a small subset while ensuring they are multi-view consistent and evenly distributed.
(3) Directly matching the dense feature map with the sampled landmarks still results in an unacceptable computational load. Therefore, we introduce a scene-specific detector that can effectively detect landmarks from the extracted dense feature map. The proposed scene-specific detector can be trained in a self-supervised manner.
(4) Based on these landmarks, the camera pose can be easily estimated by feature matching and the PnP algorithm, then refined by aligning the dense feature map with the feature field.
We conducted extensive experiments to validate the proposed pipeline's effectiveness. The results indicate that our approach surpasses state-of-the-art methods in terms of localization accuracy and recall.
\section{Related Work}
\label{sec:retlatingwork}
We categorize the related works into three sections based on scene representation: structure-based (feature matching-based) methods, neural-based methods, and radiance field-based methods.
\subsection{Structure-Based Methods}
Structure-based methods \cite{sattler2012improving,sarlin2019coarse,panek2022meshloc,kim2023ep2p,do2022learning} have demonstrated remarkable success in recent years. These approaches typically involve several key steps: feature extraction \cite{lowe1999sift,detone2018superpoint}, feature matching \cite{sarlin2020superglue,lindenberger2023lightglue, giang2024learning}, and pose estimation \cite{lepetit2009epnp} inside a RANSAC \cite{fischler1981random} loop. Active Search \cite{sattler2012improving} directly matches 2D features to the 3D point cloud to obtain 2D-3D correspondences. HLoc \cite{sarlin2019coarse} employs image retrieval method \cite{arandjelovic2016netvlad}
to achieve coarse-to-fine localization. 
This pipeline can integrate different feature detection methods \cite{detone2018superpoint, revaud2019r2d2, potje2024xfeat} and feature matching approaches \cite{sarlin2020superglue, wang2025mad} 
to improve localization performance.
Since sparse feature matching struggles with weak-texture challenges, LoFTR \cite{sun2021loftr} adopts a detector-free manner and produces semi-dense matches to improve accuracy and robustness. 
Structure-based methods make good use of scene geometry information, providing high accuracy.

\subsection{Neural-Based Methods}

Neural-based methods can be divided into absolute pose regression (APR) and scene coordinate regression (SCR), which use neural networks to represent the target scene implicitly. APR methods \cite{kendall2015posenet, chen2022dfnet, shavit2022camera, moreau2022coordinet} predict the absolute pose from a single query image. Despite their simplicity and efficiency, APR methods are often limited in accuracy and generalization in large-scale environments. SCR methods do not directly predict camera pose but instead infer the dense scene coordinates \cite{valentin2015exploiting, sattler2018benchmarking, brachmann2018learning, revaud2024sacreg, xie2023ofvl, do2022learning, tang2023neumap}. Benefiting from dense correspondences, SCR methods generally outperform structure-based methods in indoor scenes. DSAC* \cite{brachmann2021visual} and ACE \cite{brachmann2023ace} have greatly reduced the mapping time and storage cost of SCR methods. GLACE \cite{wang2024glace} improves the scalability of SCR methods to large-scale scenes by introducing co-visibility and feature diffusion, enhancing accuracy, and avoiding overfitting without relying on 3D models or depth maps. Although SCR methods can achieve high accuracy in small indoor scenes, their performance is still limited in large-scale outdoor scenes. Unlike SCR methods, our method achieves high accuracy in both indoor and outdoor scenes.

\subsection{Radiance Field-Based Methods}
Radiance field-based localization methods have recently become an active research area, driven by the impressive scene representation capabilities demonstrated by NeRF \cite{mildenhall2021nerf} and Gaussian Splatting \cite{kerbl20233d,huang20242d}.

Some methods leverage their excellent novel view synthesis capabilities for pose refinement \cite{yen2021inerf,lin2023parallel,liu2023nerf,zhao2024pnerfloc} or data augmentation \cite{moreau2022lens,martin2021nerf}. Among these, inverse rendering has emerged as a key approach for camera localization. iNeRF \cite{yen2021inerf} is the first method to directly optimize camera pose via inverse rendering, inspiring the development of subsequent methods. For example, PNeRFLoc \cite{zhao2024pnerfloc} combines 2D-3D feature matching for initial pose estimation with novel view synthesis for pose refinement, and NeRFMatch \cite{zhou2024nerfmatch} utilizes NeRF's internal features for precise 2D-3D matching and then optimizes camera poses by minimizing photometric error. Instead of photometric alignment, CROSSFIRE \cite{moreau2023crossfire} leverages dense local features from volumetric rendering to improve robustness. NeuraLoc \cite{zhai2025neuraloc} learns complementary features to establish accurate 2D-3D correspondences. In contrast to the above methods, Lens \cite{moreau2022lens} uses NeRF to synthesize additional views to expand the training dataset.
Compared with NeRF, Gaussian-based localization methods offer a more efficient alternative by leveraging 3DGS's competitive rendering quality and speed. 6DGS \cite{Bortolon20246dgs} has attracted attention by directly estimating 6DoF camera poses from a 3DGS model without requiring a prior pose. Several concurrent methods \cite{sidorov2024gsplatloc, cheng2024logs, botashev2024gsloc, zhai2025splatloc, liu2025gscpr} to ours have also made significant advances. GSplatLoc \cite{sidorov2024gsplatloc} integrates dense keypoint descriptors into 3DGS to get a coarse pose and then optimizes it using a photometric warping loss. LoGS \cite{cheng2024logs} first estimates the initial pose via image retrieval and local feature matching with a PnP solver, followed by refinement through analysis-by-synthesis. GS-CPR \cite{liu2025gscpr} leverages the 3DGS model to render high-quality images for matching to enhance the localization accuracy of neural-based methods.

In contrast to previous radiance field-based methods, our approach is a full localization pipeline that introduces a novel sparse-to-dense paradigm. The initial pose is efficiently estimated using the sampled scene landmarks and the scene-specific detector. Then, the feature field provided by Feature Gaussian is used to further refine the camera pose. Our method produces accurate pose estimation and is robust to illumination changes and weak-texture regions.

\section{Method}

\begin{figure}
  \centering
  \includegraphics[width=0.85\linewidth]{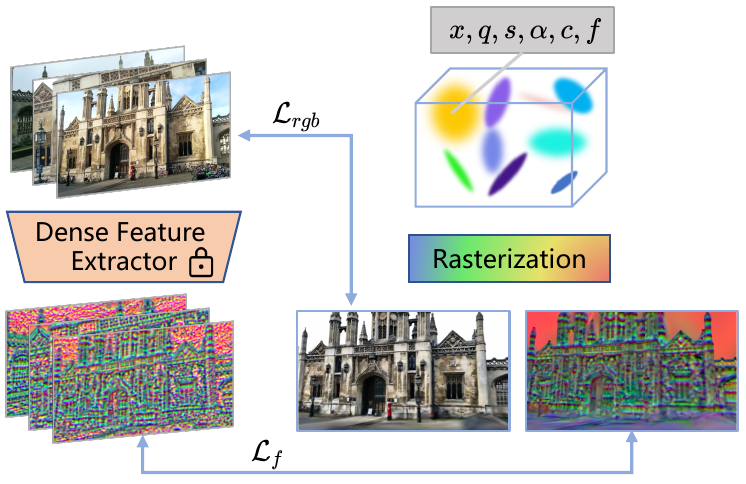}
  \caption{\textbf{Feature Gaussian} is trained by optimizing the radiance field loss $\mathcal{L}_{rgb}$ and feature field loss $\mathcal{L}_{f}$ jointly.}
  \label{fig:train featuregs}
\end{figure}

Our method is a full localization pipeline that utilizes Feature Gaussian as the scene representation. In this section, we introduce our method in detail. 
In \cref{sec:training}, we describe how to train Feature Gaussian, then in \cref{sec:sample strategy} and \cref{sec:detector}, we detail our matching-oriented sampling strategy and scene-specific detector, respectively. Finally, in \cref{sec:loc pipeline}, we present our novel sparse-to-dense camera relocalization pipeline.

\subsection{Feature Gaussian Training}
\label{sec:training}
Our scene representation consists of original Gaussian primitives augmented with a feature field. The trainable attributes of $i$-th Gaussian primitive include center $x_i$, rotation $q_i$, scale $s_i$, opacity $\alpha_i$, color $c_i$ and feature $f_i$, denoted as $\Theta_i = \{ x_i, q_i, s_i, \alpha_i, c_i, f_i \}$.
Our training process references Feature 3DGS \cite{zhou2024feature}, jointly optimizing the radiance field and the feature field, as illustrated in  \cref{fig:train featuregs}. Theoretically, our pipeline can be applied to any 3DGS variants with explicit primitives. 

We initialize the Gaussian primitives using SfM point clouds. 
For clarity and conciseness, we denote \( F_t(I) \in \mathbb{R}^{D\times H'\times W'} \) as the dense feature map extracted from the training image \( I \in \mathbb{R}^{3\times H\times W} \),  where $D$ is the dimension of the local feature.
The ground truth feature map $F_t(I)$ can be obtained using a general local feature extractor, such as SuperPoint \cite{detone2018superpoint}.
The Gaussian radiance field employs the alpha-blending technique to rasterize color attribute \( c \) into the rendered RGB image \( \hat{I} \). The same rasterize method is applied to feature attribute $f$ to render the feature map $\hat{F_s}$.

The overall training loss $\mathcal{L}$ is combined with radiance field loss $\mathcal{L}_{rgb}$ and feature field loss $\mathcal{L}_{f}$. The radiance field loss $\mathcal{L}_{rgb}$ consists of L1 loss $\mathcal{L}_1$ and D-SSIM loss $\mathcal{L}_{D-SSIM}$ between the training image $I$ and the rendered image $\hat{I}$, while the feature field loss $\mathcal{L}_{f}$ calculate L1 distance between the ground truth feature map $F_t(I)$ and the rendered feature map $\hat{F_s}$: 

\begin{equation}
    \mathcal{L}_{rgb} = (1-\lambda)\mathcal{L}_1(I, \hat{I}) + \lambda \mathcal{L}_{D-SSIM}(I, \hat{I}),
\end{equation}
\begin{equation}
    \mathcal{L}_{f} = \mathcal{L}_1(F_t(I), \hat{F_s}),
\end{equation}
\begin{equation}
    \mathcal{L} = \alpha \mathcal{L}_{rgb} + \beta \mathcal{L}_{f}.
\end{equation}
In practice, we set the weight hyperparameters $\lambda=0.2$, $\alpha=1.0$, and $\beta=1.0$ for training. We denote the complete Feature Gaussian scene obtained from training as $\mathcal{G}$.

\begin{figure}
  \centering
  \includegraphics[width=1.0\linewidth]{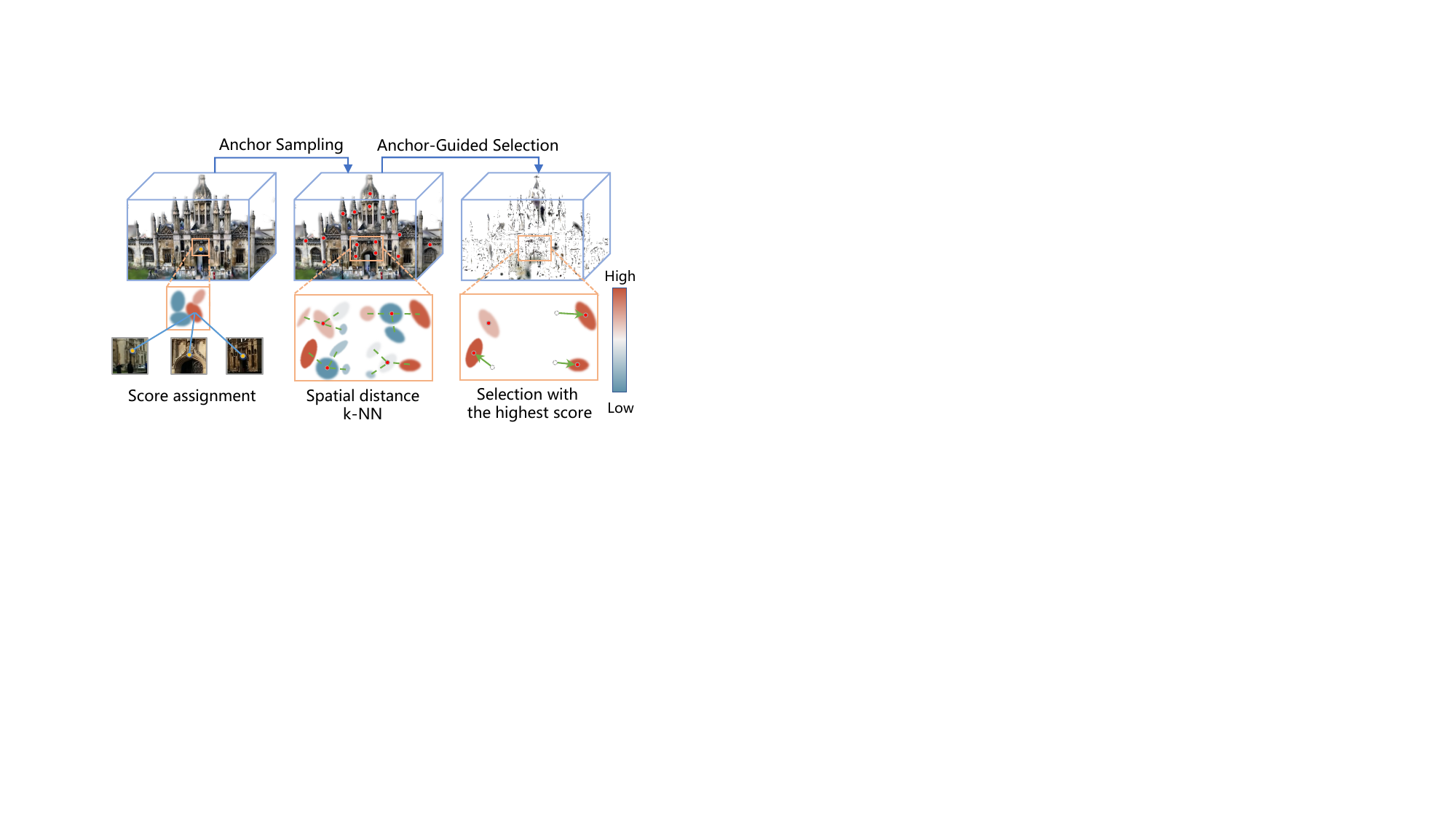}
  \caption{\textbf{Matching-Oriented Sampling.} Each Gaussian is assigned a matching score, followed by anchor sampling. For each anchor, the k nearest Gaussians are identified based on spatial distance, from which the highest-scoring Gaussian is selected.}
  \label{fig:mo_sampling}
\end{figure}

\subsection{Matching-Oriented Sampling}
\label{sec:sample strategy}
Exhaustive matching with all Gaussians in the scene is time-consuming. Besides, numerous ambiguous Gaussians may adversely affect feature matching. To address these challenges, we propose a novel matching-oriented sampling strategy, which can select Gaussians that are well-suited for matching from the millions of primitives in $\mathcal{G}$. Our objective is to ensure that the selected Gaussians are evenly distributed throughout the scene and are recognizable from various perspectives.

To quantify the quality of Gaussian, we develop a scoring strategy. Specifically, as shown in \cref{fig:mo_sampling}, for each Gaussian \( g_i \) and training image \( I \), we project the center of $g_i$ onto \( I \), yielding a 2D coordinate denoted as $(u_i, v_i)$, then we extract the corresponding 2D feature from the feature map \( F_t(I) \). The cosine similarity between the Gaussian feature $f_i$ and the extracted 2D feature is used to compute the matching score. Let \( \mathcal{V}_{i} \) denote the set of images where Gaussian $g_i$ is visible, and the final score $s(g_i)$ is obtained by averaging the matching scores across all images in \( \mathcal{V}_{i} \):
\begin{equation}
    s(g_i) = \frac{1}{\lvert \mathcal{V}_i\rvert} \sum_{I \in \mathcal{V}_i} \langle f_i, F_t(I)[u_i, v_i] \rangle.
\end{equation}
Here $F_t(I)[u_i, v_i]$ denotes the extraction of 2D feature from $F_t(I)$ at position $(u_i, v_i)$ using bilinear interpolation. A higher score indicates that the corresponding Gaussian feature is more suitable for matching.

However, selecting Gaussians solely based on their scores may lead to an uneven spatial distribution. Specifically, regions with rich textures tend to exhibit a high density of Gaussians, whereas other areas may suffer from insufficient Gaussian coverage, adversely affecting performance in those regions. To address this issue, we first employ standard downsampling techniques, such as random or farthest point sampling, to ensure a spatially uniform distribution. Specifically, we sample a fixed number of Gaussians as anchors to guide the selection process. For each anchor, we identify the $k$ nearest neighbors in terms of spatial distance and then select the Gaussian with the highest score within this neighborhood as the final result.

The number of Gaussians sampled through this method is significantly reduced compared to the original, resulting in a set of Gaussians that are uniformly distributed and highly recognizable from various perspectives. The experimental results demonstrate that sampling just a few thousand Gaussians is sufficient to achieve effective localization. We denote this set of sampled Gaussians as scene landmarks $\tilde{\mathcal{G}}$.

\begin{figure}
  \centering
  \includegraphics[width=0.7\linewidth]{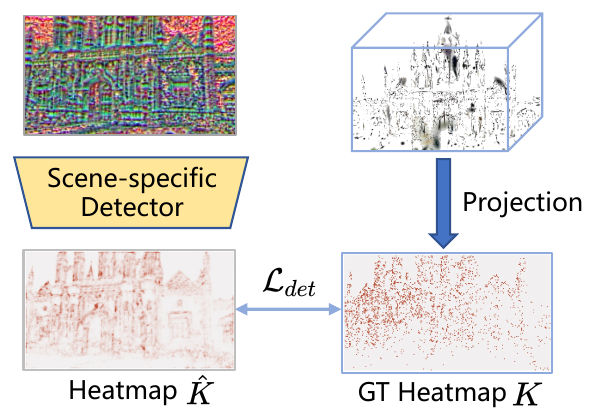}
  \caption{\textbf{Scene-Specific Detector Training.} The centers of sampled landmarks are projected onto 2D images to guide the training of the scene-specific detector.}
  \label{fig:train_ssd}
\end{figure}

\begin{figure*}
  \centering
  \includegraphics[width=1\linewidth]{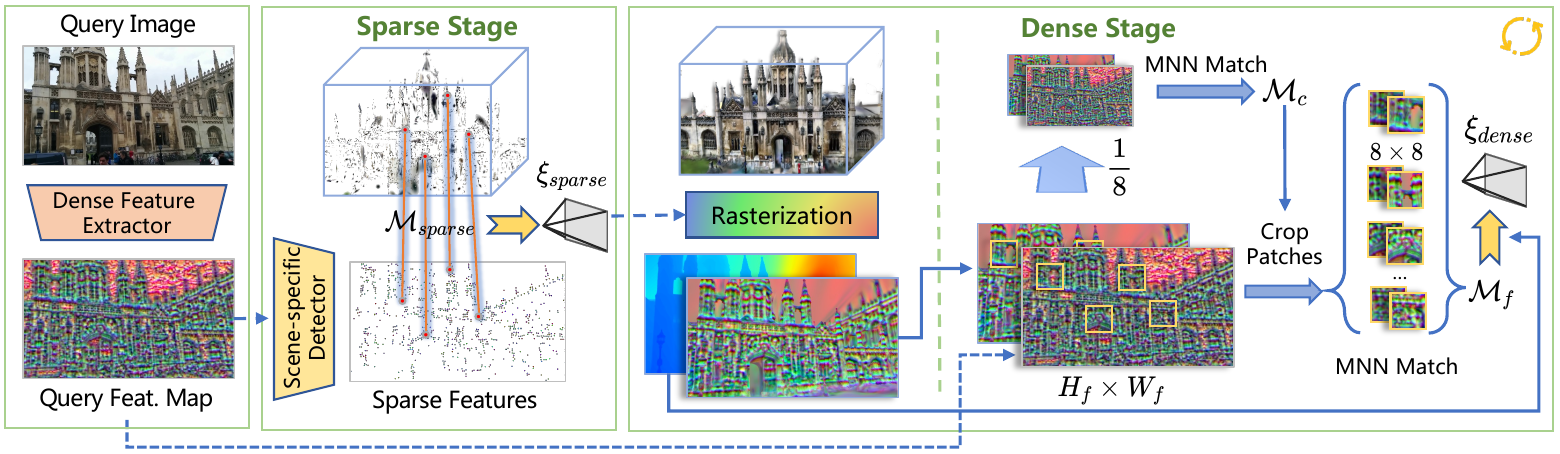}
  \caption{\textbf{Overview of the sparse-to-dense localization pipeline based on Feature Gaussian.} }
  \label{loc pipeline}
\end{figure*}

\subsection{Scene-Specific Detector}
\label{sec:detector}

Directly matching the dense feature map with sampled landmarks remains infeasible, as the dense feature map contains a large number of features, many of which are unsuitable for matching, particularly in invalid regions such as the sky. Moreover, off-the-shelf detectors such as SuperPoint \cite{detone2018superpoint} typically detect pre-defined keypoints that are scene-agnostic and therefore not well-suited for matching with the sampled landmarks in Feature Gaussian scenes.

To address this issue, we propose a scene-specific feature detector \( \mathcal{D_{\theta}} \) that processes the feature map \( F_t(I) \) and generates a heatmap \( \hat{K} \in \mathbb{R}^{1\times H' \times W'} \), indicating the probability of the 2D feature as a landmark: 

\begin{equation}
    \hat{K} = \mathcal{D_{\theta}}(F_t(I)).
\end{equation}
Specifically, our detector \( \mathcal{D_{\theta}} \) is a shallow convolution neural network (CNN), and \( \theta \) represents the network parameters.

We train \( \mathcal{D_{\theta}} \) in a self-supervised manner, as shown in \cref{fig:train_ssd}. The center of each Gaussian in \( \Tilde{\mathcal{G}} \) is projected onto the image plane, where the corresponding pixel position is set to 1 to obtain the ground truth heatmap ${K}$. We use binary cross-entropy loss to optimize detector \( \mathcal{D_{\theta}} \):

\begin{equation}
    \mathcal{L}_{\text{det}} = -K\log(\hat{K}) - (1-K)\log(1-\hat{K}).
\end{equation}

During inference, we apply non-maximum suppression (NMS) to the heatmap $\hat{K}$ to ensure that the detected keypoints are evenly distributed.

\subsection{Sparse-to-Dense Localization}
\label{sec:loc pipeline}
Thanks to the matching-oriented landmark sampling strategy and the scene-specific feature detector, we can efficiently estimate the initial pose. Furthermore, the feature field provided by Feature Gaussian supports the establishment of the sparse-to-dense localization pipeline.

The sparse-to-dense localization pipeline is illustrated in \cref{loc pipeline}. Firstly, sparse feature matching is conducted among the sampled landmarks \( \tilde{\mathcal{G}} \) and sparse local features detected by \( \mathcal{D_{\theta}} \). Based on the 2D-3D correspondences obtained from the sparse matching, the initial camera pose can be solved through the PnP algorithm. Then, a dense feature map can be rendered from the full Feature Gaussian \( \mathcal{G} \), followed by coarse-to-fine dense feature matching to refine the pose.

\textbf{Sparse Stage:} Given a query image $I_q$, we first extract a dense query feature map $F_t(I_q)$ and utilize the  \( \mathcal{D_{\theta}} \) to detect sparse local features. Then, we compute the cosine similarity between the sparse local features with all landmark features from $\tilde{\mathcal{G}}$. For each local feature, the landmark with the highest similarity is selected as a match. The 2D coordinates of the local features and the center coordinates of the corresponding landmarks are treated as 2D-3D correspondences, forming the sparse match set $\mathcal{M}_{sparse}$. Finally, the PnP algorithm with RANSAC is applied to estimate the initial pose \( \xi_{sparse} \).

\textbf{Dense Stage:} With the initial pose \( \xi_{sparse} \), we render a dense feature map $\hat{F_s}$ and a depth map $\hat{D}$ from the Feature Gaussian scene \( \mathcal{G} \). Then, we perform dense matching between $\hat{F_s}$ and $F_t(I_q)$. Inspired by LoFTR \cite{sun2021loftr}, we first conduct matching at the low resolution of $D\times H_f/8 \times W_f/8$ and then refine it at the full resolution of $D\times H_f \times W_f$. Notably, to fully leverage the dense information from the feature field, we directly render high-resolution feature maps and subsequently downsample them to low resolution via bilinear interpolation.

During matching, we first compute a correlation matrix $\mathcal{S}_c$ using cosine similarity between coarse feature maps, followed by a dual-softmax operation to obtain a probability matrix $\mathcal{P}_c$:
\begin{equation}
    \mathcal{P}_c = \text{softmax}(\frac{1}{\tau} \mathcal{S}_c)_{row} \cdot \text{softmax}(\frac{1}{\tau} \mathcal{S}_c)_{col},
\end{equation}
where $\tau$ is a temperature parameter. Then, we apply 
mutual nearest neighbor (MNN) search on $\mathcal{P}_c$ to establish coarse correspondences \( \mathcal{M}_c \).

Based on coarse matches \( \mathcal{M}_c \), we extract an $8\times 8$ patch for each coarse correspondence from the corresponding locations in the high-resolution feature map. Following the same procedure, we compute the correlation matrix $\mathcal{S}_f$ and probability matrix $\mathcal{P}_f$ on these patches and apply MNN search to obtain the refined matches \( \mathcal{M}_f \).

Finally, we lift the 2D correspondences to 3D using the rendered depth map $\hat{D}$ and solve for the pose $\xi_{dense}$ through the PnP algorithm with RANSAC. 
The dense stage allows for iterative pose refinement to achieve higher accuracy.

\section{Experiments}

\setlength{\tabcolsep}{5pt}
\begin{table*}
\centering
\begin{tabular}{@{}clcccccccc@{}}
\toprule
\multicolumn{1}{l}{}           & Method         & Chess              & Fire               & Heads              & Office             & Pumpkin            & RedKitchen         & Stairs             & Avg.↓{[}\textit{cm}/°{]}    \\ \midrule

\multirow{3}{*}{\rotatebox{90}{FM}}                             
& AS (SIFT)            & 3/0.87     & 2/1.01     & 1/0.82     & 4/1.15     & 7/1.69     & 5/1.72     & 4/1.01     & 3.71/1.18 \\ 
& HLoc (SP+SG) & 2.39/0.84 & 2.29/0.91  & 1.13/0.77  & 3.14/0.92  & 4.92/1.30  & 4.22/1.39  & 5.05/1.41 & 3.31/1.08 \\ 
& DVLAD+R2D2   & 2.56/0.88 & 2.21/0.86 & 0.98/0.75 & 3.48/1.00 & 4.79/1.28 & 4.21/1.44 & 4.60/1.27 & 3.26/1.07 \\ \midrule

\multirow{4}{*}{\rotatebox{90}{SCR}}           
& DSAC*               & \underline{0.50/0.17} & 0.78/0.29 & \underline{0.50}/0.34 & 1.19/0.35 & 1.19/0.29 & \underline{0.72/0.21} & 2.65/0.78 & \underline{1.07}/0.35 \\ 
& ACE                 & 0.55/0.18  & 0.83/0.33  & 0.53/\underline{0.33}  & \underline{1.05/0.29}  & \underline{1.06/0.22}  & 0.77/\underline{0.21}  & 2.89/0.81  & 1.10/\underline{0.34} \\ 
& NBE+SLD             & 0.6/0.18   & \underline{0.7/0.26}   & 0.6/0.35   & 1.3/0.33   & 1.5/0.33   & 0.8/\textbf{0.19}   & \underline{2.6/0.72}   & 1.16/\underline{0.34} \\
& NeuMap              & 2/0.81     & 3/1.11     & 2/1.17     & 3/0.98     & 4/1.11     & 4/1.33     & 4/1.12     & 3.14/0.95 \\ \midrule

\multirow{5}{*}{\rotatebox{90}{NeRF/GS}} 
& DFNet+NeFeS$_{50}$  & 2/0.57     & 2/0.74     & 2/1.28     & 2/0.56     & 2/0.55     & 2/0.57     & 5/1.28     & 2.43/0.79 \\
& CROSSFIRE           & 1/0.4      & 5/1.9      & 3/2.3      & 5/1.6      & 3/0.8      & 2/0.8      & 12/1.9     & 4.43/1.38 \\
& PNeRFLoc            & 2/0.80     & 2/0.88     & 1/0.83     & 3/1.05     & 6/1.51     & 5/1.54     & 32/5.73    & 7.29/1.76 \\
& NeRFMatch           & 0.95/0.30  & 1.11/0.41  & 1.34/0.92  & 3.09/0.87  & 2.21/0.60  & 1.03/0.28  & 9.26/1.74  & 2.71/0.73 \\
& \textbf{STDLoc (Ours)} &
  \textbf{0.46/0.15} &
  \textbf{0.57/0.24} &
  \textbf{0.45/0.26} &
  \textbf{0.86/0.24} &
  \textbf{0.93/0.21} &
  \textbf{0.63/0.19} &
  \textbf{1.42/0.41} &
  \textbf{0.76/0.24} \\ \bottomrule
\end{tabular}
\caption{\textbf{Localization Results on 7-Scenes.} We report the median translation errors and rotation errors for each scene. The best and second-best results are \textbf{bolded} and \underline{underlined}, respectively.}
\label{7-scenes result}
\end{table*}

\setlength{\tabcolsep}{11pt} 

\begin{table*}
\centering
\begin{tabular}{@{\hspace{10pt}}clcccccc@{\hspace{10pt}}}
\toprule
\multicolumn{1}{l}{}           & Method       & Court      & King’s          & Hospital           & Shop               & St. Mary’s        & Avg.↓{[}\textit{cm}/°{]}    \\ \midrule
\multirow{2}{*}{\rotatebox{90}{FM}}            
& AS (SIFT)     & 24/0.13 & 13/0.22 & 20/0.36 & \underline{4}/0.21 & 8/0.25 & 13.8/0.23        \\
                               
& HLoc (SP+SG) & 17.7/0.11 & \textbf{11.0}/0.20 & \underline{15.1/0.31}          & 4.2/\underline{0.20}          & \underline{7.0/0.22}          & \underline{11.0/0.21}          \\ \midrule
\multirow{4}{*}{\rotatebox{90}{SCR}}           
& DSAC*         & 33.0/0.21       & 17.9/0.31 & 21.1/0.40 & 5.2/0.24 & 15.4/0.51  & 18.5/0.33        \\
& ACE (Poker)   & 27.9/0.14       & 18.6/0.33 & 25.7/0.51 & 5.1/0.26 & 9.5/0.33   & 17.4/0.31        \\
& GLACE         & 19.0/0.12       & 18.9/0.32 & 18.0/0.42 & 4.4/0.23 & 8.4/0.29   & 13.7/0.28        \\
& NeuMap        & \textbf{6}/0.10 & 14/\underline{0.19}   & 19/0.36   & 6/0.25   & 17/0.53    & 12.4/0.29        \\ \midrule
\multirow{5}{*}{\rotatebox{90}{NeRF/GS}} 
& DFNet+NeFeS$_{50}$ & -               & 37/0.54   & 52/0.88   & 15/0.53  & 37/1.14    & 35.3/0.77        \\
& CROSSFIRE     & -          & 47/0.7          & 43/0.7             & 20/1.2             & 39/1.4            & 37.3/1.00                \\
& PNeRFLoc      & 81/0.25    & 24/0.29         & 28/0.37            & 6/0.27             & 40/0.55           & 35.8/0.35          \\
& NeRFMatch     & 19.6/\underline{0.09}       & \underline{12.5}/0.23 & 20.9/0.38 & 8.4/0.40 & 10.9/0.35  & 14.5/0.29        \\
& \textbf{STDLoc (Ours)} & \underline{15.7}/\textbf{0.06} & 15.0/\textbf{0.17}          & \textbf{11.9/0.21} & \textbf{3.0/0.13} & \textbf{4.7/0.14} & \textbf{10.1/0.14}
 \\ \bottomrule
\end{tabular}
\caption{\textbf{Localization Results on Cambridge Landmarks.} Median translation errors and rotation errors are reported, with the best results in \textbf{bold} and the second best \underline{underlined}.}
\label{cam result}
\end{table*}

\setlength{\tabcolsep}{6pt}

In this section, we first present the localization results of our pipeline on indoor and outdoor datasets in \cref{sec: loc benchmark}. Next, we conduct a detailed ablation study in \cref{sec: ablation} to evaluate the effectiveness of our matching-oriented sampling strategy, scene-specific detector, and overall localization pipeline. Finally, we provide a qualitative analysis of the localization results in \cref{sec:qual ana}.

\textbf{Datasets:} We choose the 7-Scenes dataset \cite{shotton2013scene} and the Cambridge Landmarks dataset \cite{kendall2015posenet} to evaluate our pipeline, which is widely used in visual localization benchmarks. The 7-Scenes dataset consists of 7 indoor scene sequences captured by a handheld camera, covering rich-texture and weak-texture situations. The Cambridge Landmarks dataset consists of 5 outdoor scenes captured by mobile phones, including challenges such as dynamic object occlusion, illumination changes, and motion blur. 

\textbf{Training Details:} 
Our training parameters follow the original 3DGS \cite{kerbl20233d}, attached with a learning rate of 0.001 for the feature field according to Feature 3DGS \cite{zhou2024feature}. We adopt SuperPoint \cite{detone2018superpoint} as the default feature extractor. Each scene is trained for 30,000 steps. When training on Cambridge Landmarks, we use an off-the-shelf model \cite{cheng2021mask2former} to mask out dynamic objects and the sky to reduce distractions. CLAHE histogram equalization algorithm \cite{zuiderveld1994contrast} is used to cope with illumination changes.
We set the number of sampled anchors to 16,384 and the number of nearest neighbors to 32. The scene-specific detector is a 4-layer CNN with SiLU activation, trained for 30,000 steps with a learning rate of 0.001 and an additional cosine decay.
All scenes are trained on one RTX4090 GPU, and each scene consumes about 90 minutes for Feature Gaussian training and 30 minutes for scene-specific detector training.

\textbf{Localization Details:} 
The radius of non-maximum suppression (NMS) is set to 4 pixels, and we sample 2,048 keypoints for sparse matching. 
In the dense stage, the longer side of the feature maps is set to 640 for the fine level and 80 for the coarse level. We use PoseLib \cite{PoseLib} as our default pose solver. We perform 4 iterations on 7-Scenes and 1 iteration on Cambridge Landmarks.

\subsection{Relocalization Benchmark}
\label{sec: loc benchmark}
We compare STDLoc with SOTA localization methods to demonstrate our competitive performance. The methods include structure-based methods: AS \cite{sattler2012improving}, HLoc (SP+SG) \cite{sarlin2019coarse, detone2018superpoint, sarlin2020superglue}, DVLAD+R2D2 \cite{torii201524dvlad, revaud2019r2d2}; regression-based methods: DSAC* \cite{brachmann2021dsacstar}, ACE \cite{brachmann2023ace}, NBE+SLD \cite{do2022learning}, NeuMap \cite{tang2023neumap}, GLACE \cite{wang2024glace}; and radiance field-based methods: NeFeS$_{50}$ \cite{chen2024neural}, CROSSFIRE \cite{moreau2023crossfire}, PNeRFLoc \cite{zhao2024pnerfloc},
NeRFMatch \cite{zhou2024nerfmatch}. In the HLoc and DVLAD+R2D2 experiments, we used the default setting to retrieve the top 10 images.

\begin{table}
\centering
\begin{tabular}{@{\hspace{10pt}}l@{\hspace{20pt}}c@{\hspace{20pt}}c@{\hspace{10pt}}}
\toprule
Method      & 5\textit{cm} 5°↑ & 2\textit{cm} 2°↑ \\ \midrule
HLoc (SP+SG) & 95.7\%                       & \underline{84.5\%}                       \\
DSAC*       & \underline{97.8\%}                       & 80.7\%                       \\
ACE         & 97.1\%                        & 83.3\%                        \\
NeRFMatch   & 78.2\%                        & -      \\
\textbf{STDLoc (Ours)}        & \textbf{99.1\%}              & \textbf{90.9\%}              \\ \bottomrule
\end{tabular}
\caption{\textbf{Average Recall on 7-Scenes.} We report the recall rate at thresholds of 5\textit{cm} 5° and 2\textit{cm} 2°.}
\label{7scenes recall}
\end{table}
\textbf{Indoor Localization:} We report the median translation and rotation errors on the 7-Scenes dataset in \cref{7-scenes result}. 
Our proposed method, STDLoc, improves localization accuracy and achieves the best performance on the 7-Scenes dataset. STDLoc reduces both translation and rotation errors by approximately 30\% compared to the best-performing SCR-based method ACE \cite{brachmann2023ace} and DSAC* \cite{brachmann2021dsacstar}. It is worth mentioning that STDLoc also achieves the highest recall rate among existing approaches, with 99.1\% at 5\textit{cm},5° and 90.9\% at 2\textit{cm},2°, as shown in \cref{7scenes recall}.

\textbf{Outdoor Localization:} 
We report the median translation and rotation errors on the Cambridge Landmarks dataset in \cref{cam result}.
Unlike indoor scenes, where the SCR-based method is significantly better than the FM-based method, the results in outdoor scenes show the opposite trend. The existing SOTA radiance field-based methods perform worse than the other two methods. However, STDLoc consistently outperforms all methods mentioned in the table regarding rotation accuracy. It also demonstrates competitive translation accuracy in scenes of Hospital, Shop, and St. Mary's. Due to insufficient 3D Gaussian reconstruction, the accuracy is slightly lower than HLoc and NeuMap in the other two scenes. In terms of the average metrics across all scenes, STDLoc outperforms all current SOTA methods.

\subsection{Ablation Study}
\label{sec: ablation}
\textbf{Matching-Oriented Sampling Strategy:}
The sparse stage emphasizes achieving high recall rates, so we report the 5\textit{m}, 10° recall metrics on the Cambridge Landmarks dataset, demonstrating the effectiveness of our matching-oriented sampling strategy (M.O.). We evaluate our sampling strategy by integrating it with farthest point sampling (FPS) and random sampling (RS).
In \cref{sample ablation}, the comparison between \ding{172} and \ding{174} shows that FPS performs worse than RS due to its failure to consider the distribution density of Gaussian, whereas areas with high Gaussian density typically contain more information. However, the comparisons between \ding{172} and \ding{173} as well as \ding{174} and \ding{175} demonstrate that regardless of whether FPS or RS is used, our matching-oriented sampling strategy significantly improves the quality of the sampled landmarks.

\begin{figure}
  \centering
  \includegraphics[width=0.8\linewidth]{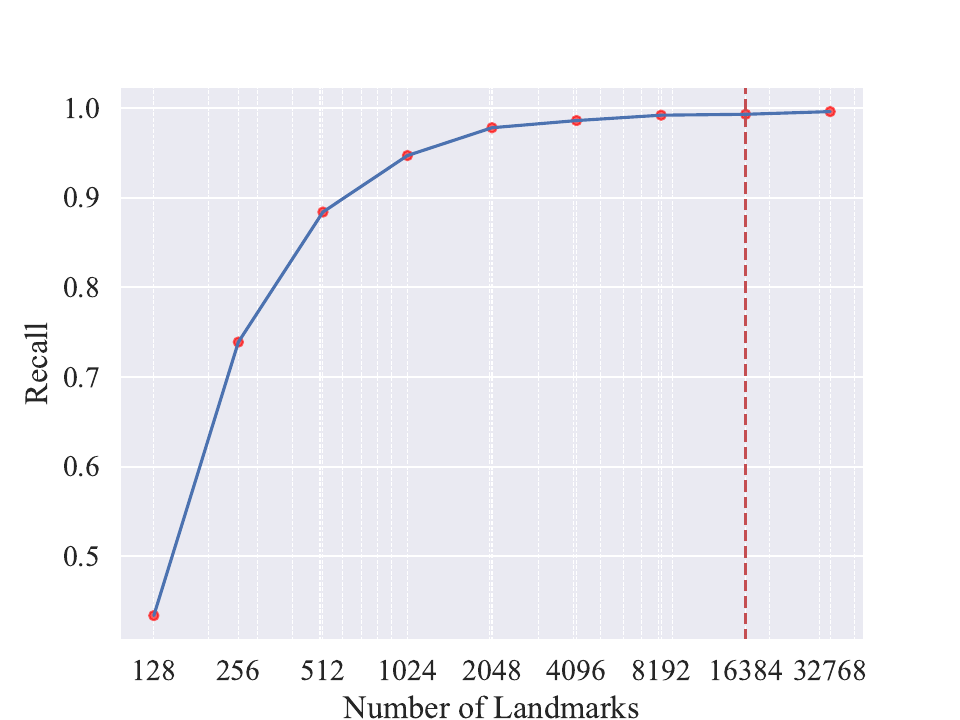}
  \caption{\textbf{Recall vs. Number of Landmarks.} We report the 5\textit{m} 10° recall on Court with varying numbers of landmarks.}
  \label{fig:recall_sample}
\end{figure}
\cref{fig:recall_sample} shows the recall trend versus the number of sampled landmarks. Combined with the detector, our sampling strategy maintains a high recall rate even at low sampling rates. With only 1,024 sampled landmarks, the recall rage still exceeds 90\%. In the experiment, we set the default sampling number to 16,384 to achieve saturated recall.

\begin{figure}
  \centering
  \includegraphics[width=0.9\linewidth]{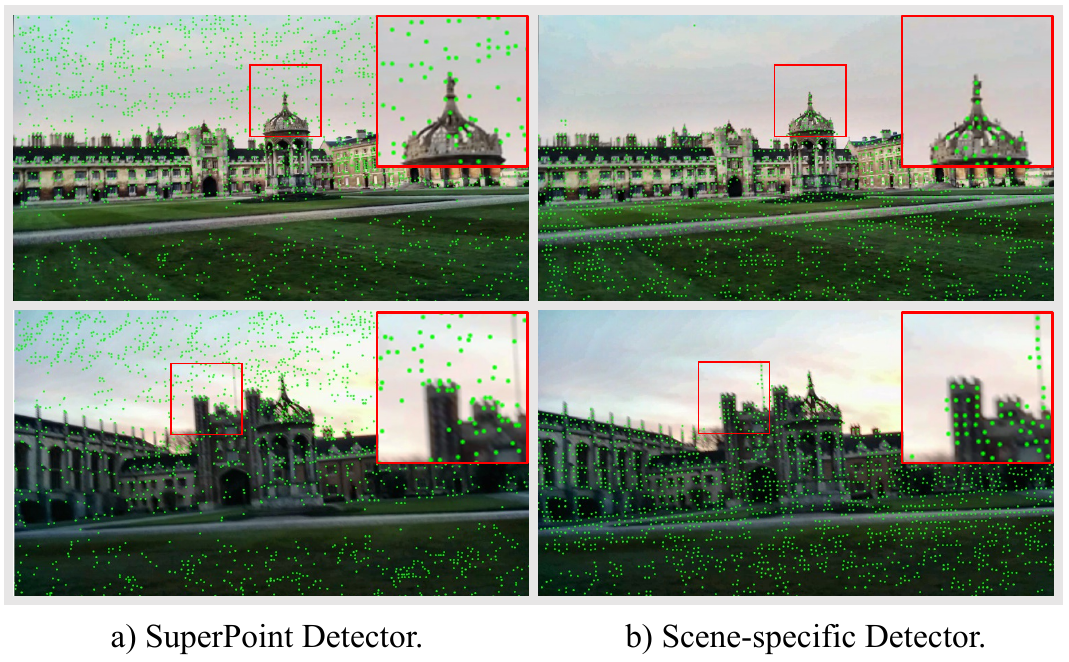}
  \caption{\textbf{Qualitative Comparison of Detectors.} We present detection results from the SuperPoint and our scene-specific detector, processed with NMS and limited to 2048 points.}
  \label{detector vis}
\end{figure}

\textbf{Scene-Specific Detector:}
We qualitatively compare our scene-specific detector (S.S.D.) with the SuperPoint detector in \cref{sample ablation}. 
The comparisons between \ding{174} and \ding{176} as well as \ding{175} and \ding{177} demonstrate that our S.S.D. effectively extracts more suitable features from the feature map for matching with sampled landmarks. We visualize the detection results in \cref{detector vis}. Our detector captures more points on buildings while fewer points in the sky, aligning more closely with the actual Gaussian distribution and yielding a higher recall. 

\begin{table}
\centering
\begin{tabular}{@{\hspace{6pt}}c|@{\hspace{10pt}}c@{\hspace{10pt}}c@{\hspace{10pt}}cc@{\hspace{6pt}}}
\toprule
No. & Sampling   & Detector & 5\textit{m} 10°↑   & 2\textit{m} 5°↑       \\ \midrule
\ding{172} & FPS      & SuperPoint       & 57.9\%   & 47.1\%       \\
\ding{173} & FPS+M.O. & SuperPoint       & 90.5\%    & 86.5\%       \\
\ding{174} & RS       & SuperPoint       & 95.0\%   & 89.7\%       \\
\ding{175} & RS+M.O.  & SuperPoint       & 98.2\%   & 96.5\%       \\
\ding{176} & RS       & S.S.D.     & 99.0\%   & 97.6\%       \\
\ding{177} & RS+M.O.  & S.S.D.     & \textbf{99.6\%} & \textbf{99.1\%} \\ \bottomrule
\end{tabular}
\caption{\textbf{Ablation Study on Sampling Strategy and Detector.} 
We report 5\textit{m} 10° and 2\textit{m} 5° recall on Cambridge Landmarks. }
\label{sample ablation}
\end{table}

\begin{table}
\centering
\begin{tabular}{@{\hspace{6pt}}lcc@{\hspace{6pt}}}
\toprule
Stage & Err.↓{[}\textit{cm}/°{]}            & 5\textit{m} 10°↑              \\ \midrule
Image Retrieval               & 586/7.9          & 48.2\%         \\
Sparse               & 13.8/0.21          & \textbf{99.6\%}         \\
Sparse + Dense (RGB)               & 15.3/0.3          & 99.4\%         \\
Sparse + Dense (Feat.)              & \textbf{10.1/0.14} & 99.4\% \\ \bottomrule
\end{tabular}
\caption{\textbf{Ablation Study on Localization Pipeline.} Average median error and recall are reported on Cambridge Landmarks.}
\label{tab:pipeline abla}
\end{table}

\begin{table}
\centering
\begin{tabular}{@{\hspace{10pt}}c@{\hspace{30pt}}c@{\hspace{40pt}}c@{\hspace{10pt}}}
\toprule
Gaussian   & Feature    &  Err.↓{[}\textit{cm}/°{]} \\ \midrule
3DGS & SuperPoint & 10.1/0.14     \\
2DGS & SuperPoint & 10.4/0.14     \\
3DGS & R2D2       & 10.1/0.15     \\
2DGS & R2D2 & 10.8/0.18     \\
\bottomrule
\end{tabular}
\caption{\textbf{Comparison of Different Gaussians and Features.} The average median error on Cambridge Landmarks is reported.}
\label{tab:Versatility}
\end{table}

\textbf{Pipeline Effectiveness:}
In \cref{tab:pipeline abla}, we report the average median error and recall for different stages of our pipeline and image retrieval method \cite{arandjelovic2016netvlad} on Cambridge Landmarks. The term ``Sparse" refers to the sparse stage, while ``Dense (Feat.)" indicates rendering the feature map directly, and ``Dense (RGB)" means rendering RGB images first, then extracting the feature map from the rendered image. Our sparse stage achieves significantly higher accuracy and recall than the image retrieval method. The Dense (Feat.) stage further enhances localization accuracy over the sparse stage. 
However, the Dense (RGB) stage performs worse, which can be attributed to the rendered RGB images being of low quality due to noise factors such as illumination changes in the training phase.
The dense stage exhibits a slight loss of recall, which can be attributed to the occlusion of floaters that prevent rendering a complete feature map and RGB image.

\textbf{Flexiblility:}
Our pipeline is capable of adapting to various explicit Gaussian representations and features. In \cref{tab:Versatility}, we conduct experiments on the Cambridge Landmarks dataset with 2DGS \cite{huang20242d} and R2D2 \cite{revaud2019r2d2} feature. The results demonstrate that our pipeline performs effectively across various combinations of Gaussians and features.

\textbf{Running Time:} We evaluate the running time of various modules of STDLoc on the Cambridge Landmarks dataset using a single NVIDIA RTX 4090 GPU. 
With one iteration for the dense stage, STDLoc achieves an inference speed of approximately 7 FPS.
The detailed time-consuming for each module can be found in the appendix \cref{tab:time}.

\subsection{Qualitative Analysis}
\label{sec:qual ana}

\begin{figure}
  \centering
  \includegraphics[width=0.9\linewidth]{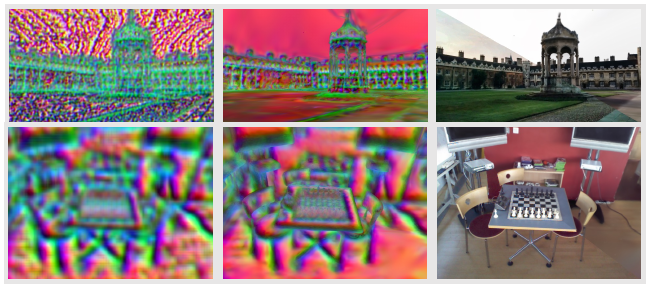}
  \caption{\textbf{Visualization Results.} We show the query feature map (left), rendered feature map (middle), and stitched image (right) of the query and rendered images. }
  \label{fig:vis_res}
\end{figure}

We present some visualization results in \cref{fig:vis_res} to provide a clearer understanding of our localization process. Our method effectively renders high-quality feature maps for dense matching and high-quality RGB images based on localization results. Additional visualizations, including failure case studies, can be found in the appendix \cref{supp:qualitative}.

\section{Conclusion}
This paper proposes a novel sparse-to-dense camera relocalization method that leverages Feature Gaussian as the scene representation. 
Based on this scene representation, we propose a new matching-oriented sampling strategy and a scene-specific detector to facilitate efficient and robust sparse matching to obtain the initial pose. 
Then, the localization accuracy is improved by aligning the dense feature map to the trained feature field through dense matching.
Our proposed method can produce accurate pose estimation and is robust to illumination-change and weak-texture situations. Experimental results demonstrate that our method outperforms the current SOTA methods.

\textbf{Limitations and Future Work:} 
Due to the training limitations of Feature Gaussian, our method cannot be extended to very large scenes. However, this may be addressed through techniques such as Level of Detail (LoD) on 3DGS or chunked training. Additionally, the floating artifacts in 3DGS can adversely affect localization. Reducing these artifacts is an important research direction for 3DGS. We believe that advancements in 3DGS can be easily integrated into our framework, further improving localization performance.


{
    \small
    \bibliographystyle{ieeenat_fullname}
    \bibliography{main}
}

\setcounter{section}{0} 
\setcounter{figure}{0}  
\setcounter{table}{0}   

\maketitlesupplementary

\renewcommand\thesection{\Alph{section}} 
\renewcommand\thesubsection{\thesection.\arabic{subsection}} 
\renewcommand\thefigure{\Alph{figure}} 
\renewcommand\thetable{\Alph{table}} 

\crefname{section}{Sec.}{Secs.}
\Crefname{section}{Section}{Sections}
\Crefname{table}{Table}{Tables}
\crefname{table}{Tab.}{Tabs.}

\newcommand{\tabnohref}[1]{Tab.~{\color{red}#1}} 
\newcommand{\fignohref}[1]{Fig.~{\color{red}#1}} 
\newcommand{\secnohref}[1]{Sec.~{\color{red}#1}} 
\newcommand{\cnohref}[1]{[{\color{green}#1}]} 
\newcommand{\linenohref}[1]{Line~{\color{red}#1}}

\section{Feature Gaussian Training}
\label{sec:training_details}
Our training strategy references Feature 3DGS \cite{zhou2024feature}. To adapt the feature field for the localization task and improve robustness, we make the following modifications:

\begin{enumerate}
    \item Thanks to the development of the CUDA accelerated rasterization tool gsplat \cite{ye2024gsplatopensourcelibrarygaussian}, we can efficiently render feature maps while preserving the original feature dimensions, eliminating the need for the speed-up module proposed in Feature 3DGS for upsampling feature channels after rasterization. Specifically, the feature \( f \) stored in the Gaussian primitive \( g \) has the same dimension \( D \) as the feature map \( F_t(I) \) extracted using the general local feature extractor. This also enables direct matching between the 2D query features and the 3D features of the Gaussian primitives.
    
    \item The rendering process for the radiance field of Feature Gaussian is based on the alpha blending rasterization method \cite{alphablending1995}. Let \( \mathcal{N} \) denote the set of Gaussians associated with a pixel, sorted in front-to-back order. The pixel color \( C \) is computed by blending the color \( c \) of Gaussians as follows:
    \begin{equation}
        C = \sum_{i \in \mathcal{N}} c_i \alpha_i T_i,
    \end{equation}
    where \( T_i \) is the transmittance factor accounting for the accumulated opacity $\alpha$ of all preceding Gaussians, defined as:
    \begin{equation}
        T_i = \prod_{j=1}^{i-1} (1 - \alpha_j).
    \end{equation}
    This alpha blending approach is also applied to render the feature field. However, due to the vector triangle inequality, directly accumulating features is unsuitable. To address this, we introduce L2 normalization into the alpha blending formula. Specifically, we normalize the Gaussian feature \( f \) before rasterization to mitigate the influence of feature magnitude, and we further normalize the accumulated features after rasterization. The final rendered feature \( F_s \) is therefore expressed as:
    \begin{equation}
        F_s = \text{norm}\left(\sum_{i=1}^{n} \text{norm}(f_i) \alpha_i T_i\right),
    \end{equation}
    where \( \text{norm}(\cdot) \) denotes the L2 normalization operation. This two-step normalization process ensures stability and robustness in the feature field training and rendering.
\end{enumerate}

\begin{table}
    \centering
    \begin{tabular}{@{\hspace{10pt}}l@{\hspace{50pt}}c@{\hspace{10pt}}}
    \toprule
    Module                   & Time (ms) \\ \midrule
    Feature Extraction       & 3.7       \\
    S.S.D.                   & 6.4       \\
    Sparse Matching          & 17.4      \\
    Pose Estimation (Sparse) & 15.8      \\
    Rasterization            & 23        \\
    Dense Matching           & 13.2      \\
    Pose Estimation (Dense)  & 72.8      \\ \midrule
    Total                    & 152.3     \\ \bottomrule
    \end{tabular}
    \caption{\textbf{Detailed Time Consumption Analysis.}}
    \label{tab:time}
\end{table}

\section{Matching-Oriented Sampling Algorithm}
The algorithm of the matching-oriented sampling strategy is illustrated in \cref{alg:sampling}. 

\begin{figure*}
  \centering
  \includegraphics[width=0.75\linewidth]{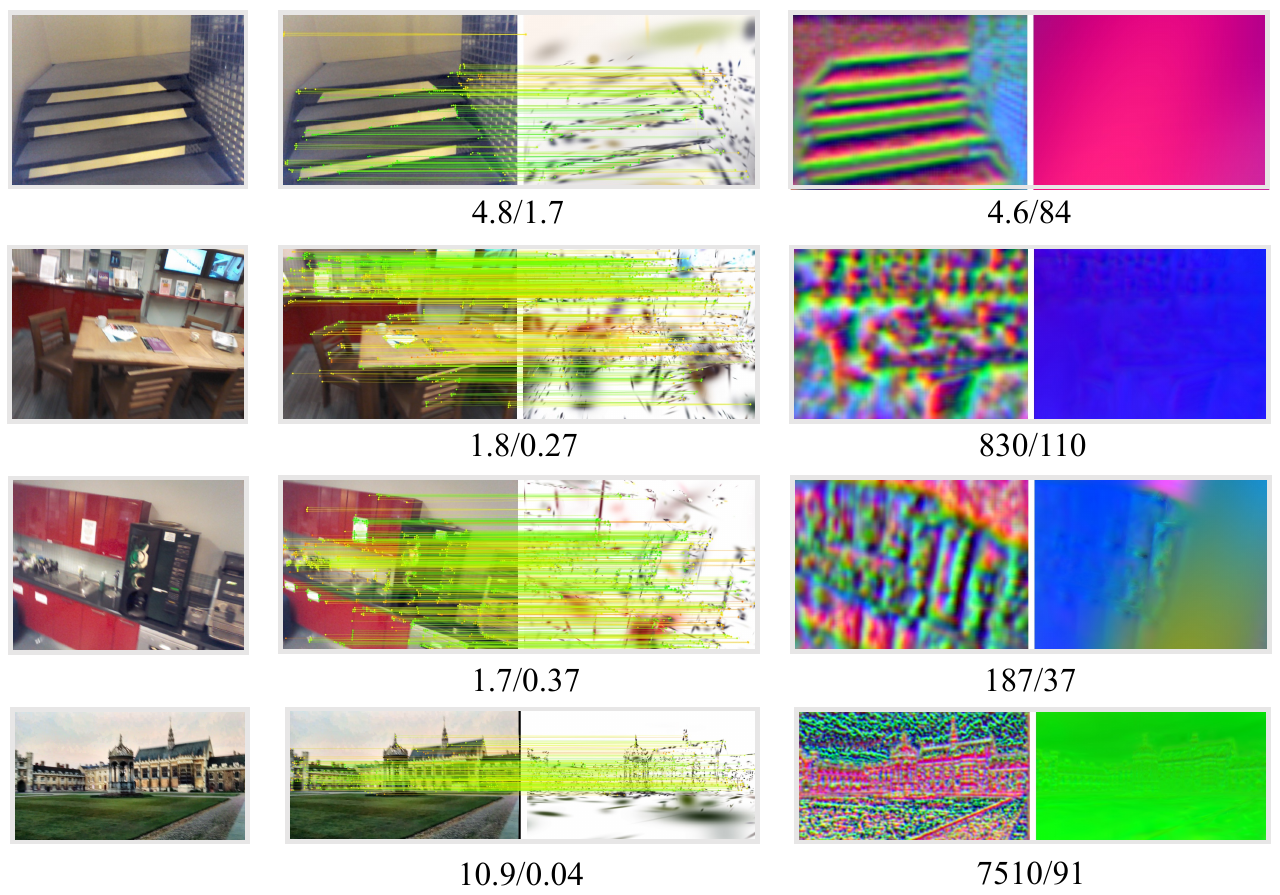}
  \caption{
  \textbf{Failure Cases Visualization.} Visualization results of some examples where localization is successful in the sparse stage but failed in the dense stage. The translation error (\textit{cm}) and rotation error (°) are indicated below the corresponding stage.
  }
  \label{fig:failure_case}
\end{figure*}

\section{Qualitative Analysis}
\label{supp:qualitative}

\begin{figure}
  \centering
  \includegraphics[width=0.9\linewidth]{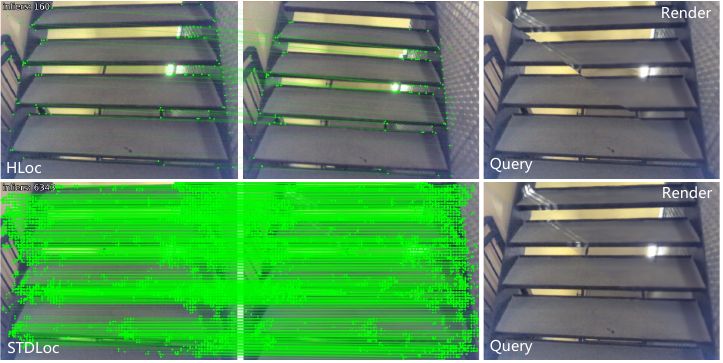}
  \caption{\textbf{Comparison with HLoc in Weak Texture Scenario.} }
  \label{fig:supp_textureless}
\end{figure}

\begin{figure}
  \centering
  \includegraphics[width=0.9\linewidth]{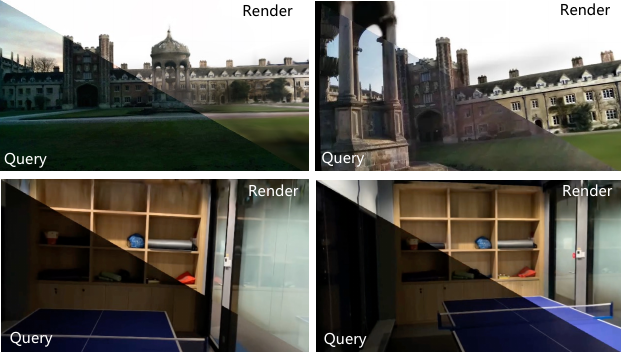}
  \caption{\textbf{Localization Results in Illumination Changes Scenarios.} }
  \label{fig:supp_illumination}
\end{figure}

\textbf{Challenging Cases.}
In \cref{fig:supp_textureless} \cref{fig:supp_illumination}, we present the localization results of STDLoc in challenging scenarios involving weak texture and varying illumination conditions. For illumination changes, we demonstrate sample cases from both the Cambridge Landmarks dataset and real scene, where STDLoc achieves precise localization results. In the weak texture scenario, we provide a comparative analysis with HLoc in the Stairs scenario from the 7-Scenes dataset. STDLoc extracts denser matches, enabling more accurate pose estimation.

\textbf{Failure Cases.} 
As shown in \cref{tab:pipeline abla}, the recall of (5\textit{m}, 10°) in the dense stage is slightly lower than that in the sparse stage. The localization results for these failure cases are illustrated in \cref{fig:failure_case}. The first column is the query image, the second column is the sparse matching result between the query image and landmarks, and the third column is the dense feature map of the query image and the feature map rendered based on the sparse stage localization pose.

The sparse stage successfully provides an accurate pose estimation, but the feature map rendered in the dense stage is indistinct, leading to the failure of dense matching. This lack of distinguishability in the dense feature map is caused by floaters in the scene, which is a common issue in 3DGS \cite{kerbl20233d} scenes. Therefore, reducing floaters in the scene can be effective in minimizing these failure cases. In addition, this situation reflects the robustness of our sparse stage, which can effectively eliminate the influence of these floaters through the matching-oriented sampling strategy.

\begin{figure*}[!ht]
  \centering
  \includegraphics[width=0.93\linewidth]{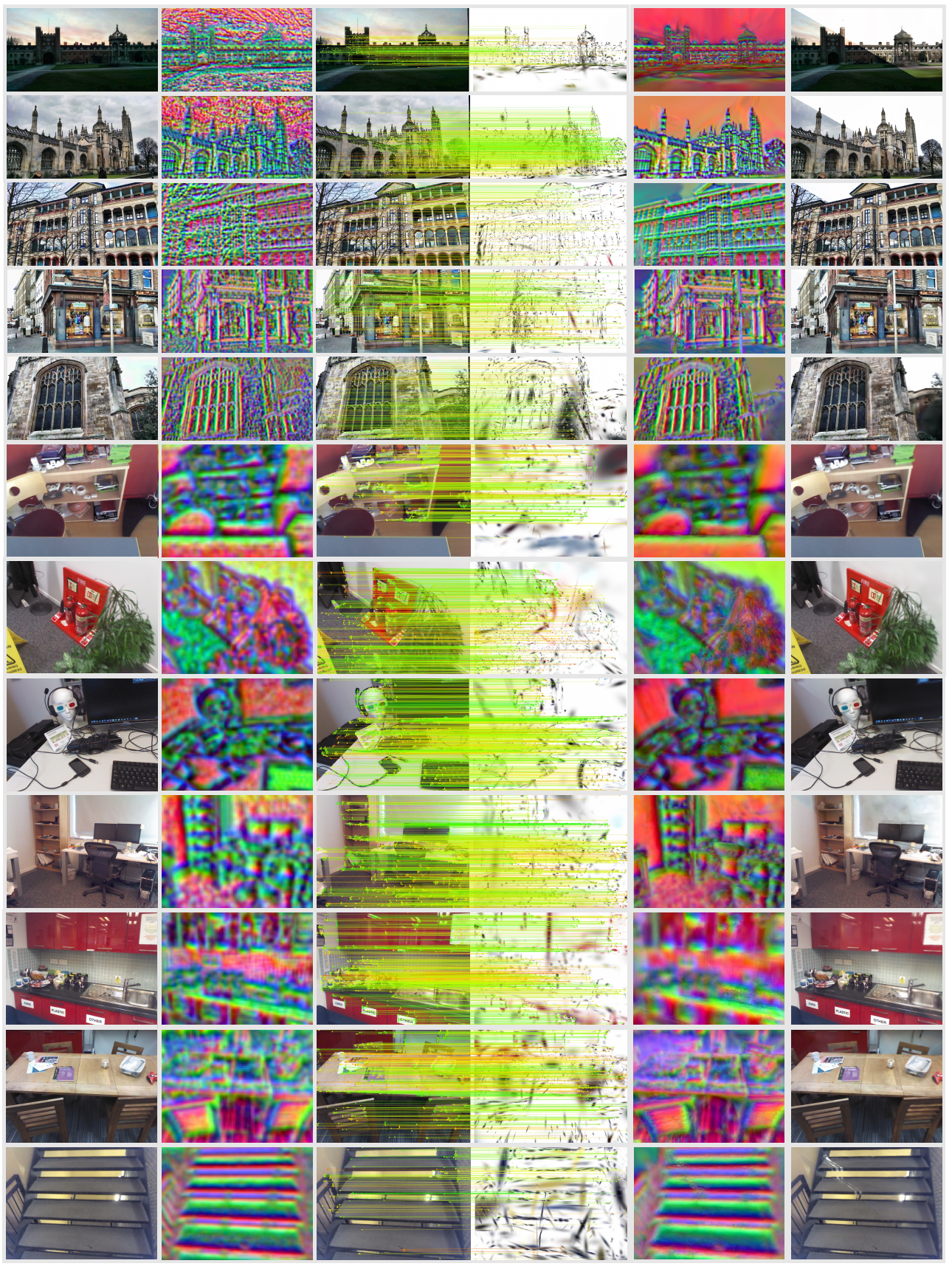}
  \caption{
 \textbf{More Visualizations.} We show all scenes on both the Cambridge Landmarks and 7-Scenes datasets.}

  \label{fig:vis_case}
\end{figure*}

\textbf{More Visualization Results.}
\cref{fig:vis_case} presents the localization visualization results across all scenes for both the Cambridge Landmarks and 7-Scenes datasets. From left to right, each column shows the query image, its corresponding dense feature map, the sparse matching result between the query image and landmarks, the rendered feature map from the final dense stage, and the stitched result of the query image with the rendered image using the final pose. The visualization of the sparse matching results is achieved by rendering Gaussian landmarks based on the pose estimated in the sparse stage, followed by drawing the matches.

The third column of the figure demonstrates that our sparse stage achieves robust 2D-3D matching results. This is attributed to our proposed matching-oriented sampling strategy and scene-specific detector. Furthermore, the second-to-last column demonstrates the capability to learn the feature field using Feature Gaussian. As shown in the last column, the rendered image aligns precisely with the query image, highlighting the high accuracy of our localization method. By leveraging the learned feature field, our approach exhibits remarkable robustness against illumination changes and weak texture.

\begin{algorithm*}
\caption{Matching-Oriented Sampling Algorithm}
\label{alg:sampling}
\begin{algorithmic}[1]
\Require{Gaussians $\mathcal{G}$, training images $\{I\}$, feature maps $\{F_t(I)\}$, anchor number $n$, nearest neighbors $k$}
\Ensure{Sampled landmarks $\tilde{\mathcal{G}}$}

\For{each Gaussian $g \in \mathcal{G}$}
    \Comment{Assign scores for each Gaussian}
    \State $f \gets \text{norm}(g.feature)$ \Comment{Normalize Gaussian feature}
    \State $\mathcal{V} \gets \{\text{The set of images where } g \text{ is visible}\}$
    \State $s \gets 0$
    \For{each Image $I \in \mathcal{V}$}
        \State $(u, v) \gets \text{Project}(g.center, I)$ \Comment{The pixel coordinates of the projected Gaussian center}
        \State $f_{\text{img}} \gets \text{norm}(\text{GridSample}(F_t(I), (u, v)))$ \Comment{Extract 2D feature using bilinear interpolation}
        \State $s \gets s + \langle f, f_{\text{img}} \rangle$
    \EndFor
    \State $g.score \gets s / |\mathcal{V}|$ \Comment{Average score across images}
\EndFor

\State $\mathcal{A} \gets \text{RandomSampling}(\mathcal{G}, n)$ \Comment{Randomly sample anchors}
\State $\tilde{\mathcal{G}} \gets \emptyset$

\For{each anchor $a \in \mathcal{A}$}
    \Comment{Anchor-guided selection}
    \State $\mathcal{N}_a \gets \text{FindkNearestNeighbors}(a, \mathcal{G}, k)$
    \State $g^* \gets \mathop{\arg\max}\limits_{g \in \mathcal{N}_a} g.score$ \Comment{Select Gaussian with the highest score among neighbors}
    \State $\tilde{\mathcal{G}} \gets \tilde{\mathcal{G}} \cup \{g^*\}$
\EndFor
\end{algorithmic}
\end{algorithm*}

\end{document}